\documentclass[letterpaper, 10 pt, conference]{ieeeconf}  

\IEEEoverridecommandlockouts                              

\usepackage{graphicx} 
\usepackage{subcaption}
\usepackage[font=footnotesize,labelfont=bf]{caption}
\usepackage{amsmath,amsthm,amssymb}
\usepackage{xcolor}
\usepackage[ruled,vlined,linesnumbered]{algorithm2e}
\usepackage{cite}
\usepackage{url}
\usepackage{comment}
\renewcommand{\tilde}{\widetilde}
\newcommand{\mD}{\mathcal{D}}
\newcommand{\mG}{\mathcal{G}}
\newcommand{\mK}{\mathcal{K}}
\newcommand{\mP}{\mathcal{P}}
\newcommand{\mU}{\mathcal{U}}
\newcommand{\mW}{\mathcal{W}}

\newcommand{\R}{\mathbb{R}}

\newcommand{\bu}{\mathbf{u}}
\newcommand{\bx}{\mathbf{x}}
\newcommand{\dd}{\mathrm{d}}
\newcommand{\tp}{\mathsf{T}}
\mathchardef\mhyphen="2D
\newcommand{\red}[1]{\textcolor{red}{#1}}

\newcommand{\norm}[1]{\left\Vert #1 \right\Vert}
\newcommand{\tuple}[1]{\langle #1 \rangle}
\newcommand{\diag}{\operatorname{diag}}
\SetKw{KwInit}{Initialize:}
\SetKw{KwInput}{Input:}
\SetKw{KwOut}{Output:}
\SetKw{KwBreak}{break}
\SetKw{KwOr}{or}

\theoremstyle{remark}
\newtheorem*{remark}{Remark}

\title{\LARGE \bf
Stackelberg Game-Theoretic Trajectory Guidance for Multi-Robot Systems with Koopman Operator  
}

\author{Yuhan Zhao and Quanyan Zhu
\thanks{Y. Zhao and Q. Zhu are with the Department of Electrical and Computer Engineering, New York University, Brooklyn, NY, 11201, USA. Email:{\{yhzhao, qz494\}@nyu.edu}.}
}

\begin{document}

\maketitle
\thispagestyle{empty}
\pagestyle{empty}

\begin{abstract}
Guided trajectory planning involves a leader robot strategically directing a follower robot to collaboratively reach a designated destination. However, this task becomes notably challenging when the leader lacks complete knowledge of the follower's decision-making model. There is a need for learning-based methods to effectively design the cooperative plan. To this end, we develop a Stackelberg game-theoretic approach based on the Koopman operator to address the challenge. We first formulate the guided trajectory planning problem through the lens of a dynamic Stackelberg game. We then leverage Koopman operator theory to acquire a learning-based linear system model that approximates the follower's feedback dynamics. Based on this learned model, the leader devises a collision-free trajectory to guide the follower using receding horizon planning.
We use simulations to elaborate on the effectiveness of our approach in generating learning models that accurately predict the follower's multi-step behavior when compared to alternative learning techniques. Moreover, our approach successfully accomplishes the guidance task and notably reduces the leader's planning time to nearly half when contrasted with the model-based baseline method\footnote{The simulation codes are available at \url{https://github.com/yuhan16/Stackelberg-Koopman-Learning}.}.
\end{abstract}

\section{Introduction} \label{sec:intro}
As a special kind of cooperation in multi-robot systems, guided cooperation allows multiple robots of heterogeneous capabilities to work together with leader-follower/mentor-apprentice types of interactions and jointly achieve the task objective. It has been studied in many fields including human-robot collaboration \cite{bo2016human,cacace2022combining}, multi-agent path planning \cite{ding2010multi,bibuli2012guidance}, collective transportation \cite{machado2016multi,fu2022leader,koung2021cooperative}. In guided cooperation, a more resourceful robot can assist a less sophisticated robot in completing the task by leveraging the capabilities of the two agents. As an example, we consider navigation guidance between two unmanned ground vehicles (UGVs). One UGV has limited sensing and planning capabilities and has difficulty reaching the target destination independently. The other UGV, equipped with better sensing capabilities and planning resources, can find a collision-free trajectory to guide the first UGV to the destination.

Stackelberg games \cite{bacsar1998dynamic} provide a quantitative framework to capture such guided interactions. In the two-UGV example, the more powerful UGV acts as a leader and plans the future trajectory by anticipating the other UGV's behavior; the less resourceful UGV refers to the leader UGV's guidance and decides its next move. Their interactions can be formulated as a dynamic Stackelberg game. The resulting Stackelberg equilibrium (SE) solution provides a feasible agent-wise cooperative plan for navigation.

However, computing the SE requires the leader robot to know the follower robot's decision-making model, including the objective function and the dynamics, which can be challenging to obtain in practice. In many cases, we can only observe the past interactive trajectories. Even if the follower's model is known, model-based algorithms to compute the SE, such as mixed integer programming \cite{paruchuri2008playing}, have many computational challenges. There is a need for learning-based methods to find the SE and enable fast planning in guided cooperation tasks.

Koopman operator theory provides a data-driven method to learn a linear model of nonlinear systems from trajectories \cite{brunton2019data,bevanda2021koopman,brunton2022modern,mezic2021koopman}. It also enables efficient model predictive control algorithms using the learned linear model \cite{korda2018linear,proctor2018generalizing,brunton2016koopman}. Using the Koopman-operator approach, the leader can estimate the follower's decision-making model using a linear system to anticipate the follower's behavior. Based on the estimated model, the leader can compute an approximate SE solution to make cooperative plans.

\begin{figure}
    \centering
    \includegraphics[height=4.5cm]{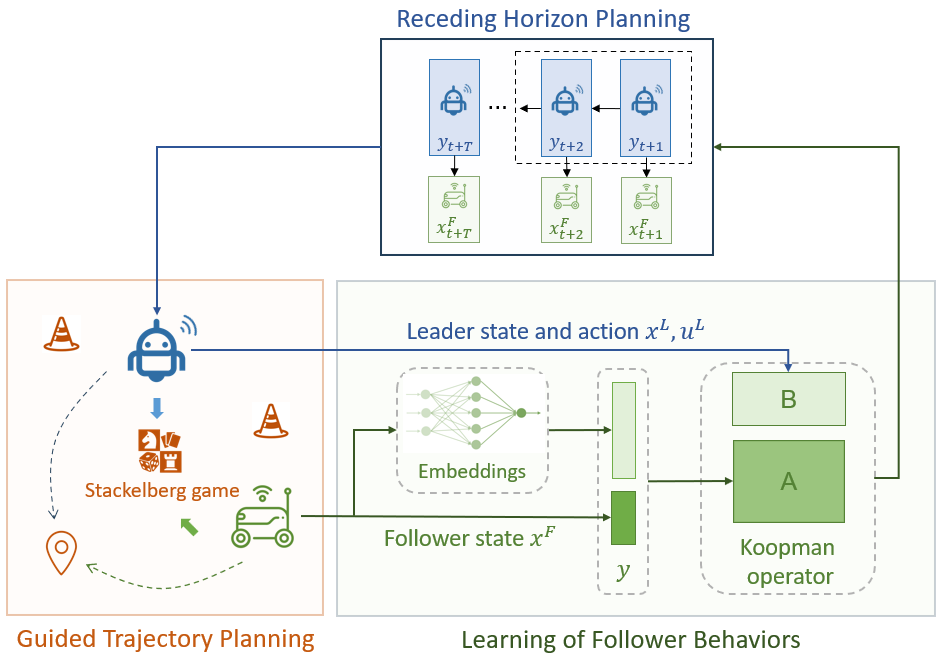}
    \captionsetup{belowskip=-15pt}
    \caption{Illustration of Stackelberg Koopman learning framework in trajectory guidance with an unknown follower robot. The guidance problem is modeled as a Stackelberg game. The leader leverages the Koopman operator to learn a linear model to predict the follower's behavior. She performs receding horizon planning to find a collision-free trajectory and guide the follower to the target destination.}
    \label{fig:intro}
\end{figure}

In this work, we develop a Stackelberg game-theoretic approach based on the Koopman operator to address the challenging multi-robot guided cooperation problem, in which the follower’s planning parameters are unknown to the leader.
We first formulate the guided cooperation as a dynamic Stackelberg game, where the follower makes \emph{myopic} decisions to interact with the leader and decides his next move. To compute the SE, the leader leverages Koopman operator theory to learn an approximate linear model of the follower's \emph{feedback} dynamics with the best response. Based on the learned model, the leader computes the approximate SE using receding horizon planning to interact with the follower to complete the guidance task. We use simulations to demonstrate that our approach can successfully complete the guidance in an environment with obstacles. Compared with another two learning methods, our approach provides effective linear models to predict the follower's multi-step behavior. Besides, our approach reduces the leader's planning time by half compared with the model-based baseline method while achieving successful guidance in different scenarios.

\section{Related Work} \label{sec:related_work}
Originating from the seminal work \cite{koopman1931hamiltonian,koopman1932dynamical}, the Koopman operator was proposed to predict the evolution of autonomous dynamical systems using an infinite dimensional linear system. To facilitate computation, finite-dimensional approximation of the infinite linear system can be found by learning the Koopman invariant subspaces or eigenfunctions \cite{takeishi2017learning,lusch2018deep,yeung2019learning}.
Recent studies in the Koopman operator have prospered its development in both theory and applications \cite{mauroy2020koopman,manzoor2023vehicular,ling2020koopman}. Various learning algorithms have been developed to estimate the Koopman operator more effectively. For example, Shi et al. in \cite{shi2022deep} have developed three deep learning algorithms to identify the Koopman embeddings for nonlinear control systems and learn the corresponding linear models for predictive control. Mamakoukas et al. in \cite{mamakoukas2021derivative} have leveraged higher-order derivatives of general nonlinear dynamics to improve the learning accuracy of the Koopman-operator-based linear representation and the robustness of the prediction under measurement noise.
In robotics, most work leverages the Koopman operator to learn an approximate linear model of the robot's dynamical system and performs model predictive control (MPC). For example, Koopman-based system identification methods for controlling soft robots with precision have been discussed in \cite{bruder2019nonlinear,bruder2020data,wang2022improved}. The work \cite{folkestad2020episodic} by Folkestad et al. has applied Koopman operator theory to learn the nonlinear dynamical systems of multirotors and achieved fast landing using MPC. Similar approaches have been applied to wheeled mobile robots to achieve system identification and trajectory tracking \cite{ren2022koopman}. However, few work has studied the application of Koopman operator in game theory and multi-robot systems. Our paper aims to bridge the gap and introduce the Koopman operator to multi-agent settings.

The Stackelberg game characterizes leader-follower cooperative interactions between agents and can be used to design and evaluate collaboration plans for guidance tasks \cite{koh2020cooperative,zhao2022stackelberg,fisac2019hierarchical,sadigh2016planning,nikolaidis2017game}. For example, Zhao and Zhu in \cite{zhao2022stackelberg} have leveraged Stackelberg games to coordinate heterogeneous robots and jointly complete multi-object rearrangement tasks. Fisac et al. in \cite{fisac2019hierarchical} has studied efficient and safe trajectory planning for autonomous vehicles in the human driver-aware environment based on Stackelberg games.
When there is incomplete information in the Stackelberg game, learning-based methods are helpful in finding the SE. Reinforcement learning has become a popular model-free approach to learning the SE through interactions \cite{zhang2020bi,meng2022learning,kononen2004asymmetric}. However, in Stackelberg game-based applications, it is pragmatic to assume that the leader knows her decision-making model but does not know the follower. It is sufficient for the leader to learn the follower's model to compute the SE and find collaboration plans. In this work, we propose an approach using Koopman operator theory to learn a follower's decision-making model and seek an approximate SE for guided trajectory planning.

\section{Stackelberg Game Formulation} \label{sec:sg}

\subsection{Trajectory Guidance as Dynamic Stackelberg Games} \label{sec:sg.formulation}
We consider a leader robot $L$ (she) guiding a follower robot $F$ (he) in the working space $\mW$ to reach the target destination while avoiding obstacles. Due to the limited sensing and planning capabilities, the follower can only make one-step (myopic) local planning to decide the next move. Therefore, he needs guidance from a more resourceful leader who can sense the entire environment and make multi-step predictions to find a collision-free trajectory.

Let $x^L_t \in \R^{n^L}$ and $x^F_t \in \R^{n^F}$ be the leader and follower's states at time $t$, including their positions. Let $u^L_t \in \mU^L$ and $u^F_t \in \mU^F$ be their controls and admissible control sets, respectively. We further denote $x_t := [x^L_t, x^F_t]$ as the joint state at time $t$ and $x^\dd$ as the common target state. During the interactive guidance, two robots first observe the joint state $x_t$. The leader determines an action $u^L_t$ and announces it to the follower. Next, the follower myopically seeks a one-step-optimal action $u^{F*}_t$ to move based on his objective function $J^F$ and the leader's action $u^L_t$. Two robots then move to the new state $x_{t+1}$ and start the next round of guidance. 
We assume $M$ obstacles in $\mW$ and define the safety constraint $c^i_j(x^i) \geq 0$, $i \in \{L, F\}, j=1,\dots, M$, to achieve obstacle avoidance.
Using the follower's responses, the leader computes $T$-step actions $\bu^{L} := \{ u^L_t \}_{t=0}^{T-1}$ that generates a collision-free trajectory to guide the follower to the target destination. The trajectory guidance problem can be formulated as a dynamic Stackelberg game $\mG$ as follows:
\begin{align}
    \min_{\bu^L \in \mU^L} \ & J^L(\bu^L, \bu^F) := \sum_{t=0}^{T-1} g^L_t(x_t, u^L_t, u^F_t) + g_T^L(x_T) \label{eq:sg.lobj} \\ 
    \text{s.t.} \quad & x^L_{t+1} = f^L(x_t, u^L_t), \quad t = 0,\dots, T-1, \label{eq:sg.ldyn} \\ 
    & c^L_j(x_t) \geq 0, \quad j = 1,\dots, M, \ t = 1,\dots, T, \label{eq:sg.lsafe} \\
    & 
        \min_{u^F_t \in \mU^F} \quad J^F(u^F_t; x_t, u^L_t) \quad t = 0,\dots, T-1, \label{eq:sg.fobj} \tag{4a} \\ 
        & \quad \text{s.t} \hspace{8mm}  x^F_{t+1} = f^F(x^F_t, u^F_t), \label{eq:sg.fdyn} \tag{4b} \\ 
        & \hspace{15mm} c^F_j(x^F_{t+1}) \geq 0, \quad j = 1,\dots, M, \label{eq:sg.fsafe} \tag{4c} \\
    & x^L_0, x^F_0 \text{ given}. \notag
\end{align}
Here, $T$ is the leader's planning horizon. $J^L$ and $\{ g^L_t \}_{t=0}^T$ in \eqref{eq:sg.lobj} denote the leader's objective and stage costs. Eq. \eqref{eq:sg.ldyn}-\eqref{eq:sg.lsafe} capture the leader's dynamics and safety constraints. The follower's myopic planning problem is given by \eqref{eq:sg.fobj}-\eqref{eq:sg.fsafe}, where $J^F$ in \eqref{eq:sg.fobj} is his objective function that takes the current state $x_t$ and the leader's action $u^L_t$ as parameters. \eqref{eq:sg.fdyn} is the follower's dynamics and \eqref{eq:sg.fsafe} ensures one-step safety for the follower. 

\begin{remark}
In this work, we consider regular obstacles that can be represented by norms. The safety constraint for obstacle $j = 1,\dots, M$ and robot $i \in \{L, F\}$ can be defined by
\begin{equation*}
    c^i_j(x^i) := \norm{ \Lambda_j \left( p(x^i) - p^O_j \right)}_l - r_j \geq 0.
\end{equation*}
Here, $p(x^i)$ is the position of robot $i$ recovered from the state $x^i$; $p^O_j$ is the position of the $j$-th obstacle center; $\Lambda_j$ and $r_j$ determine the obstacle's shape and range; $l$ denotes the $l$-norm. For example, we set $l=l_2$ for circular obstacles and $l=l_\infty$ for rectangle obstacles.
\end{remark}

We use the open-loop SE $\tuple{\bu^{L}_{\text{SE}}, \bu^F_{\text{SE}}}$ of the game $\mG$ as the agent-wise collaboration plan for the guidance task. The leader and follower take the actions in $\bu^{L}_{\text{SE}}$ and $\bu^{F}_{\text{SE}}$, respectively, to generate a collision-free trajectory $\bx := \{ x_t \}_{t=1}^{T}$ starting from the initial state $x_0 := [ x^L_0, x^F_0 ]$.

\subsubsection{Guidance in Trajectory Planning}
The guidance is an interactive process rather than unilateral instructions from the leader. The follower can make (myopic) decisions for the next move, and the leader's action serves as a reference to assist the follower in making effective local planning. Without guidance, the follower will not reach the target destination due to limited sensing and planning capabilities.

Here, the leader needs to predict the follower's behavior when computing the SE. The leader's prediction coincides with the follower's real behavior if the leader has complete information about the follower, including the follower's objective and dynamics. However, if the leader only has an estimated follower model, her prediction can deviate from the follower's real action. Thus, the follower always plans his action instead of following the leader's instructions.

\subsection{Model-Based Approach for SE}
The bilevel programming nature of the game $\mG$ challenges the computation of the SE.
We leverage reformulation techniques to obtain tractable algorithms. Note that the follower is myopic and there are $T$ inner-level problems (follower's problem) in $\mG$. 
For the $t$-th inner-level problem, we penalize the safety constraints with log barrier functions and define
\begin{equation*}
    \tilde{J}^F_t(u^F_t; x_t, u^L_t) := J^F - \frac{1}{\mu} \sum_{j=1}^M \log \left( c^F_j(f^F(x^F_t, u^F_t)) \right),
\end{equation*}
where $\mu > 0$ is the penalty parameter. Then, the $t$-th inner-level problem becomes 
\begin{equation*}
    \min_{u^F_t \in \mU^F} \quad \tilde{J}^F_t(u^F_t; x_t, u^L_t).
\end{equation*}
We use the first-order condition $\nabla_{u^F_t} \tilde{J}^F_t = 0$ to approximate the optimality condition since $\tilde{J}^F_t$ is nonconvex. Thus, the leader's problem becomes a single-agent Optimal Control Problem (OCP), denoted by $\mP_{\text{foc}}$:
\begin{equation}
\label{eq:ocp.foc} \tag{5}
\begin{split}
    \min_{\bu^L, \bu^F} \quad & J^L(\bu^L, \bu^F) \\ 
    \text{s.t.} \quad & x^L_{t+1} = f^L(x^L_t, u^L_t), \quad t=0,\dots, T-1, \\
    & c^L_i(x_t) \geq 0, \quad j=1\dots, M, \ t = 1, \dots, T. \\ 
    & \nabla_{u^F_t} \tilde{J}^F_t(u^F_t; x_t, u^L_t) = 0, \quad t=0, \dots, T-1, \\
    & x^L_0, x^F_0 \text{ given},
\end{split}
\end{equation}
which can be efficiently solved by trajectory optimization methods, such as direct collocation. We denote the solution of \eqref{eq:ocp.foc} as $\tuple{\tilde{\bu}^L_{\text{foc}}, \tilde{\bu}^F_{\text{foc}}}$, which can be used to approximate the SE $\tuple{\bu^L_{\text{SE}}, \bu^F_{\text{SE}}}$ of $\mG$ for generating cooperative plans.

\subsection{Guidance with Unknown Follower} \label{sec:sg.unknown}
The model-based approach \eqref{eq:ocp.foc} provides an approximate solution of $\mG$. However, the leader must know the follower's entire model \eqref{eq:sg.fobj}-\eqref{eq:sg.fsafe}. In many cases, the leader can only observe the follower's states and actions through interactions rather than the decision-making model. Therefore, we need learning-based methods to compute the SE of the game $\mG$.

From the follower's problem \eqref{eq:sg.fobj}-\eqref{eq:sg.fsafe}, we define follower's feedback policy $\rho^F: \R^{n^F} \times \R^{n^L} \times \mU^L \to \mU^F$ by: 
\begin{equation*}
    \rho^F(x^F_t; x^L_t, u^L_t) =  \arg\min_{u^F_t \in \mU^F} \eqref{eq:sg.fobj} \mhyphen \eqref{eq:sg.fsafe}.
\end{equation*}
Note that the feedback policy $\rho^F$ is a stationary function since the follower's problem remains the same for all $t$. The follower uses it to generate actions and update his dynamical system, yielding \emph{feedback dynamics}:
\begin{equation}
\label{eq:fb_fdyn} \tag{6}
    x_{t+1}^F = f^F \left( x^F_t, \rho^F(x^F_t; x^L_t, u^L_t) \right) := \tilde{f}^F(x_t, u^L_t).
\end{equation}
Once the leader knows the feedback policy $\rho^F$, her objective function can be written as $J^L(\bu^L, \rho(\bx, \bu^L)) := \tilde{J}^L(\bu^L)$. Her decision-making problem (i.e., the game $\mG$) can be reduced to the following OCP, denoted by $\mP_{\text{fb}}$:
\begin{equation}
\label{eq:ocp.fb} \tag{7}
\begin{split}
    \min_{\bu^L \in \mU^L} \quad & \tilde{J}^L(\bu^L) \\ 
    \text{s.t.} \quad & x^L_{t+1} = f^L(x^L_t, u^L_t), \quad t=0,\dots, T-1, \\
    & x^F_{t+1} = \tilde{f}^F(x_t, u^L_t), \quad t=0,\dots, T-1, \\
    & c^L_j(x_t) \geq 0, \quad j=1\dots, M, \ t = 1, \dots, T. \\ 
    & x^L_0, x^F_0 \text{ given},
\end{split}
\end{equation}
The optimal solution of \eqref{eq:ocp.fb}, denoted by $\bu^L_{\text{fb}}$, coincides with the SE of the leader's part $\bu^L_{\text{SE}}$. In fact, the follower's response trajectory, denoted by $\bu^F_{\text{fb}}$,  coincides with $\bu^F_{\text{SE}}$. In other words, the leader can solve \eqref{eq:ocp.fb} and use the optimal solution to guide the follower, which essentially captures the collaborative planning through the SE approach.
Note that the leader in \eqref{eq:ocp.fb} does not predict the follower's action because the follower's optimal action generated by $\rho^F$ is already incorporated in the feedback dynamics $\tilde{f}^F$. 
The Koopman operator provides a data-driven method to learn a linear model of nonlinear systems. In the next section, we present a Koopman-based approach to enable learning-based methods to approximate the SE.

\section{Koopman Operator for Feedback Dynamics} \label{sec:kp}
\subsection{Basics of Koopman Operator}
We consider a discrete nonlinear dynamical system 
\begin{equation*}
    x_{t+1} = f(x_t), \quad \text{where } f: \R^n \to \R^n.
\end{equation*}
The Koopman operator $\mK$ is an infinite-dimensional linear operator that acts on the embedding function $g: \R^n \to \R^q$ to measure the state evolution along the system trajectory:
\begin{equation*}
    \mK g(x_t) = g \circ f(x_t) = g(x_{t+1}), \quad  t=0,1,\dots.
\end{equation*}
The Koopman operator has also been extended to dynamical systems with controls \cite{proctor2018generalizing}. For a system $x_{t+1} = f(x_t, u_t)$ with $f: \R^n \times \R^m \to \R^n$, the Koopman operator acting on the extended embedding function $g: \R^n \times \R^m \to \R^q$ yields
\begin{equation*}
    \mK g(x_t, u_t) = g \circ f(x_t, u_t) = g(x_{t+1}, u_{t+1}).
\end{equation*}
Using the embedding function $g$, the Koopman operator lifts the state and control spaces to a higher dimensional embedding space to obtain a new linear system that is easier to handle.

Note that Koopman operator $\mK$ is an infinite-dimensional operator. We aim to find a finite matrix representation $K \in \R^{q\times q}$ of $\mK$ such that $g(x_{t+1}, u_{t+1}) \approx K g(x_t, u_t)$. 
To obtain $K$, we split the embedding function $g$ into two parts $g = [g_x(x,u)^\tp \ g_u(x,u)^\tp]^\tp$ and further assume that $g_x$ is only related to the state, i.e., $g_x(x) := g_x(x,u)$. Then, we obtain
\begin{equation*}
    \begin{bmatrix} g_x(x_{t+1}) \\  g_u(x_{t+1}, u_{t+1}) \end{bmatrix} = \begin{bmatrix} K_{xx} & K_{xu} \\ K_{ux} & K_{uu} \end{bmatrix} \begin{bmatrix} g_x(x_t) \\  g_u(x_t, u_t) \end{bmatrix},
\end{equation*}
which gives a linear system in terms of embedding functions: 
\begin{equation} 
\label{eq:kp_sys} \tag{8}
    g_x(x_{t+1}) = K_{xx} g_x(x_t) + K_{xu} g_{u}(x_t, u_t).
\end{equation}
Many parameterization methods on $g_u(x, u)$ are proposed to recover the control $u_t$. The linear parameterization is the most common, which sets $g_u(x,u) = u$. The advantage of linear parameterization is that it preserves the control $u$ in the learned system, which makes it more convenient to perform MPC. In this work, we adopt linear parameterization. 

Given a $S$-step state-action trajectory $\{ \hat{x}_t, \hat{u}_t, \hat{x}_{t+1} \}_{t=0}^{S-1}$, we first compute the embedding state $\hat{g}_t := g_{x}(\hat{x}_t)$. The Koopman operator in \eqref{eq:kp_sys} is computed by optimizing the $S$-step prediction loss:
\begin{equation*}
    K^*_{xx}, K^*_{xu} = \arg\min_{K_1, K_2} \sum_{t=0}^S \gamma^s \norm{ \hat{g}_{t+1} - K_1 \hat{g}_t - K_2 \hat{u}_t}^2_2,
\end{equation*}
where $\gamma \in (0,1]$ is the weight decay factor. The use of $S$-step loss contributes to the accuracy of long-term prediction.

\subsection{Learning Feedback Dynamics Using Koopman Operator} \label{sec:kp.learn}
We leverage Koopman operator theory to learn the follower's feedback dynamics $\tilde{f}^F$ in \eqref{eq:fb_fdyn}. Note that we need to learn both the Koopman operator $K$ and the embedding function $g_x$. We parameterize $g_x$ with some parameter $\theta$ (denoting $h_\theta$ as the parameterized function) and further encode the state itself to $g_x$, which gives $g_x(x) = [x^\tp \ h_\theta(x)^\tp]^\tp$.
Since $\tilde{f}^F$ outputs the follower's state $x^F$, we denote $w^L:=[x^L, u^L]$ as the augmented variable and let $y_t = g_x(x^F_t)$. $x^F_t$ can be recovered by $x^F_t = C y_t$ where $C = [I_{n^F} \ 0]$. Then, we obtain the following linear system
\begin{equation} 
\label{eq:kp_linear} \tag{9}
\begin{split}
    y_{t+1} &= K_{xx} y_{t} + K_{xu} w^L := Ay_t + B w^L_t \\ 
    & \hspace{29.5mm} :=A y_t + B_1 x^L_t + B_2 u^L_t, \\ 
    x^F_t &= C y_t.
\end{split}
\end{equation}
Here, we split $B := [B1 \ B2]$ for clarity. We can leverage \eqref{eq:kp_linear} to predict the follower's future state given the leader's state-action pair $\tuple{x^L_t, u^L_t}$ and the initial $x^F_0$.

Using the interactive trajectory (length $S$) data set $\mD = \{ \{\hat{x}^{L}_{t,i}, \hat{x}^F_{t,i}, \hat{u}^L_{t,i}, \hat{u}^F_{t,i}, \hat{x}^L_{t+1,i}, \hat{x}^F_{t+1,i}\}_{t=0}^S \}_{i=1}^N$, the leader learns $A, B$ and the embedding functions $g_x$ by minimizing the loss 
\begin{equation*}
    L(\theta, A, B) = \frac{1}{N} \sum_{i=1}^N \sum_{t=0}^S \gamma^s \| g_x(\hat{x}^F_{i,t}) - A g_x(\hat{x}^F_{i,t}) - B \hat{w}^L_{i,t} \|^2_2,
\end{equation*}
which can be optimized by Stochastic Gradient Descent (SGD). Then, the leader uses the learned model in lieu of the follower's feedback dynamics in \eqref{eq:ocp.fb} and solves the following problem, denoted as $\mP_{\text{kp}}$, to approximate the solution of \eqref{eq:ocp.fb}:
\begin{equation}
\label{eq:ocp.kp} \tag{10}
\begin{split}
    \min_{\bu^L \in \mU^L} \  & \tilde{J}^L(\bu^L) \\ 
    \text{s.t.} \quad  & x^L_{t+1} = f^L(x_t, u^L_t), \quad  t = 0,\dots, T-1, \\
    & y_{t+1} = A y_{t} + B_1 x^L_t + B_2 u^L_t, \ \ t = 0,\dots, T-1, \\
    & c^L_j(x^L_t) \geq 0, \quad j = 1,\dots, M, \ t = 1,\dots, T, \\ 
    & x^L_0, y_0(x^F_0) \text{ given}.
\end{split}
\end{equation}
We denote the solution of \eqref{eq:ocp.kp} as $\tilde{\bu}^L_{\text{kp}}$, which approximates $\bu^L_{\text{fb}}$ and hence $\bu^L_{\text{SE}}$. With the learned model, the leader uses $\tilde{\bu}^L_{\text{kp}}$ to guide the follower. The follower's response under $\tilde{\bu}^L_{\text{kp}}$, denoted by $\tilde{\bu}^F_{\text{kp}}$, together with $\tilde{\bu}^L_{\text{kp}}$, constitutes an approximation of the SE and a collaboration plan.

\subsection{Scenario-based Receding Horizon Planning} \label{sec:kp.rc}
We use receding horizon planning to perform the guidance. The leader periodically updates the guidance trajectory based on the learned model to overcome the approximation errors in Koopman learning. We note that guidance becomes less meaningful if the leader is too far away from the follower. Therefore, we design a scenario-based planning mechanism by introducing a guidance threshold. When the distance between the leader and the follower exceeds the threshold, the leader will return to the follower. Otherwise, the leader moves toward the destination to guide the follower. The follower's objective remains unchanged in the guidance.
The scenario-based planning mechanism can be achieved by switching the leader's objective function, which is summarized in Alg.~\ref{alg:1}.

\begin{algorithm}
\KwInit Initial state $x^L_0, x^F_0$, destination $x^\dd$, scenario threshold $\lambda$ \;
Leader learns the Koopman operator by minimizing $S$-step loss \;
\For{$t = 0,1,2,\dots$}{
    Leader computes embedding state $y_t$ using $x^F_t$ \;
    \eIf{$\operatorname{dist}(x^L_t - x^F_t) > \lambda$}{
        Leader sets objective 1 to go to the follower \;
    }{
        Leader sets objective 2 to head to destination \;
    }
    $\bu^{L*} \gets$ Leader solves OCP \eqref{eq:ocp.kp} \;
    Leader applies $u^{L*}_{0}$ \;
    Follower observes $x_t, u^{L*}_0$ and computes $u^{F*}_0$ \;
    Leader and follower update dynamics $x^L_{t+1}, x^F_{t+1}$ \;
    $x^L_t \gets x^L_{t+1}$, $x^F_t \gets x^F_{t+1}$ \;
    \uIf{Follower reaches destination}{
        \KwBreak \;
    }
}
\caption{Receding horizon planning for strategic trajectory guidance.}
\label{alg:1}
\end{algorithm}

\section{Simulations and Evaluations} \label{sec:exp}
\subsection{Simulation Settings}
\begin{figure*}[h]
    \captionsetup[subfigure]{justification=centering}
    \begin{subfigure}[b]{0.24\textwidth}
        \centering
        \includegraphics[height=3.5cm]{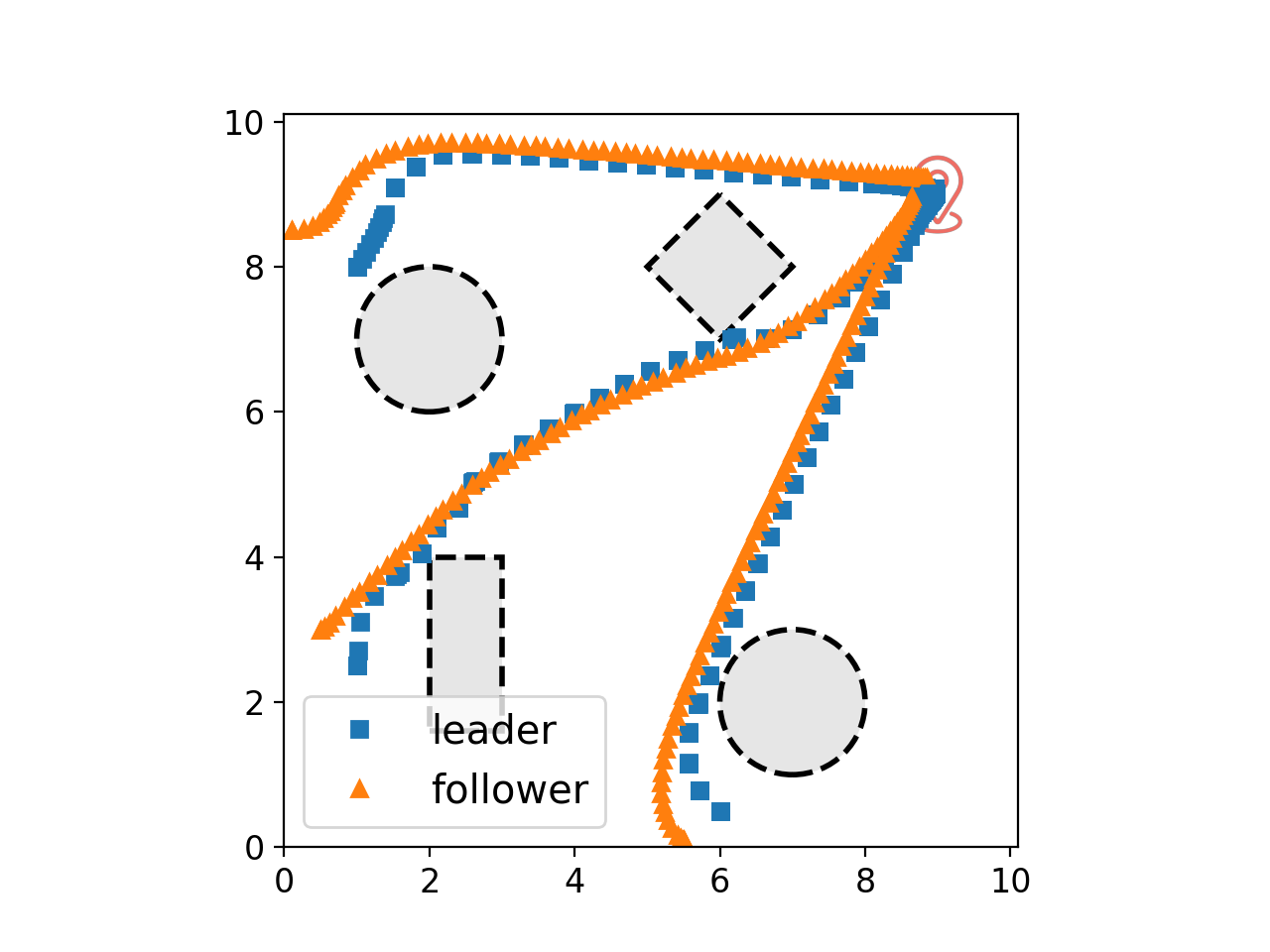}
        \caption{Model-based OCP (baseline).}
        \label{fig:rc.nonlin}
    \end{subfigure}
    \captionsetup[subfigure]{justification=centering}
    \begin{subfigure}[b]{0.24\textwidth}
        \centering
        \includegraphics[height=3.5cm]{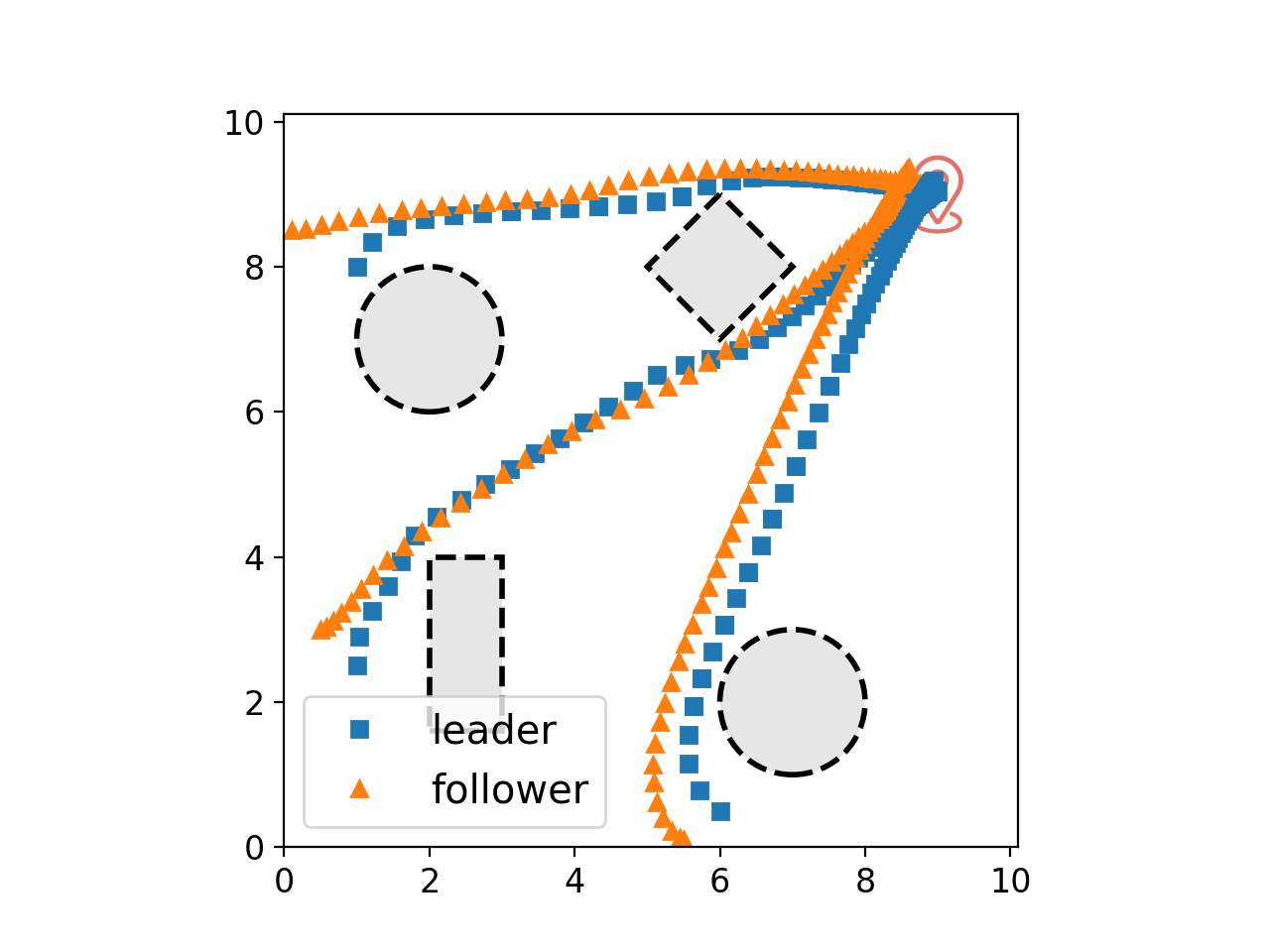}
        \caption{Our approach.} 
        \label{fig:rc.kp}
    \end{subfigure}
    \captionsetup[subfigure]{justification=centering}
    \begin{subfigure}[b]{0.24\textwidth}
        \centering
        \includegraphics[height=3.5cm]{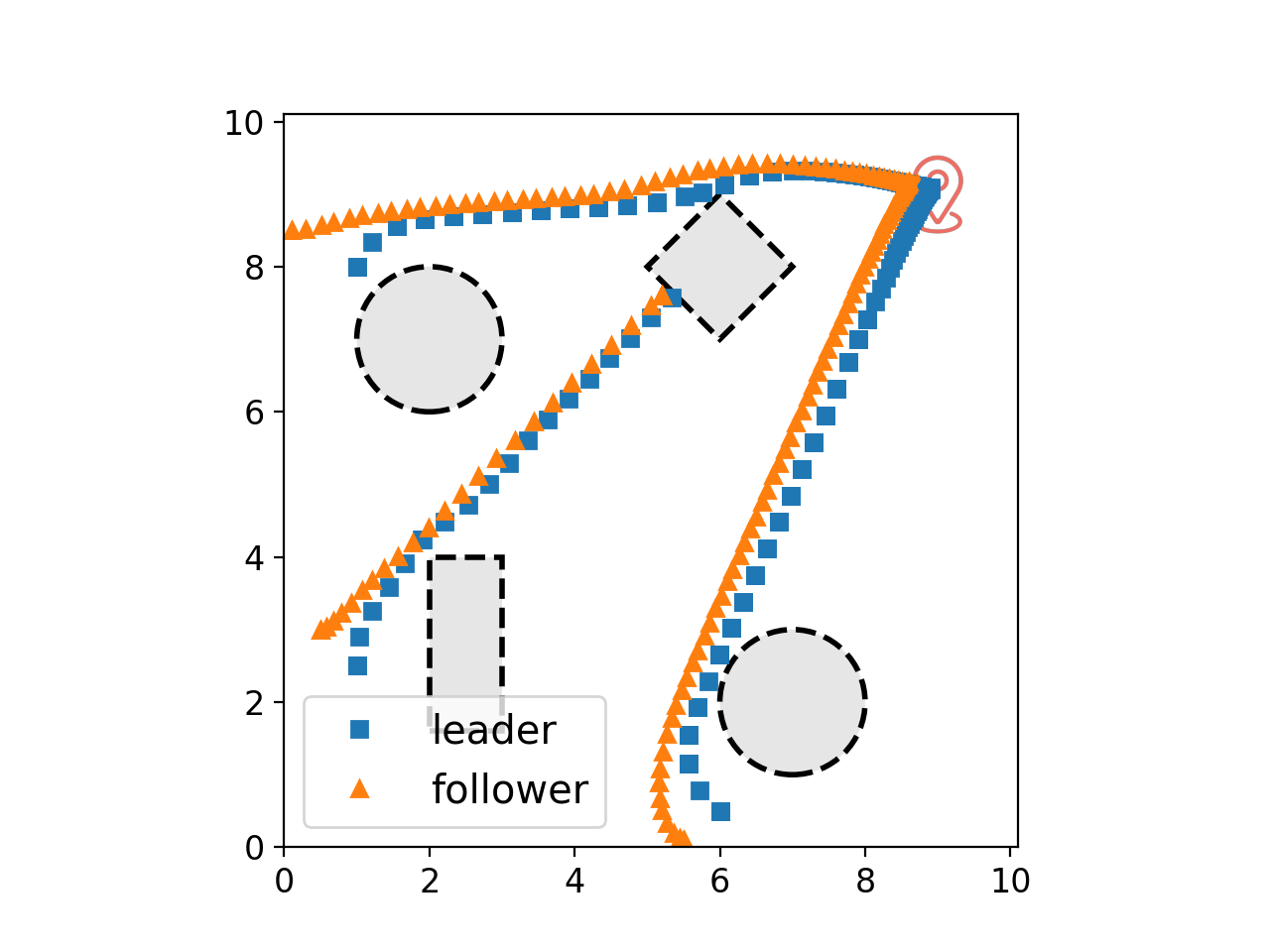}
        \caption{\texttt{nn}-approach.} 
        \label{fig:rc.nn}
    \end{subfigure}
    \captionsetup[subfigure]{justification=centering}
    \begin{subfigure}[b]{0.24\textwidth}
        \centering
        \includegraphics[height=3.5cm]{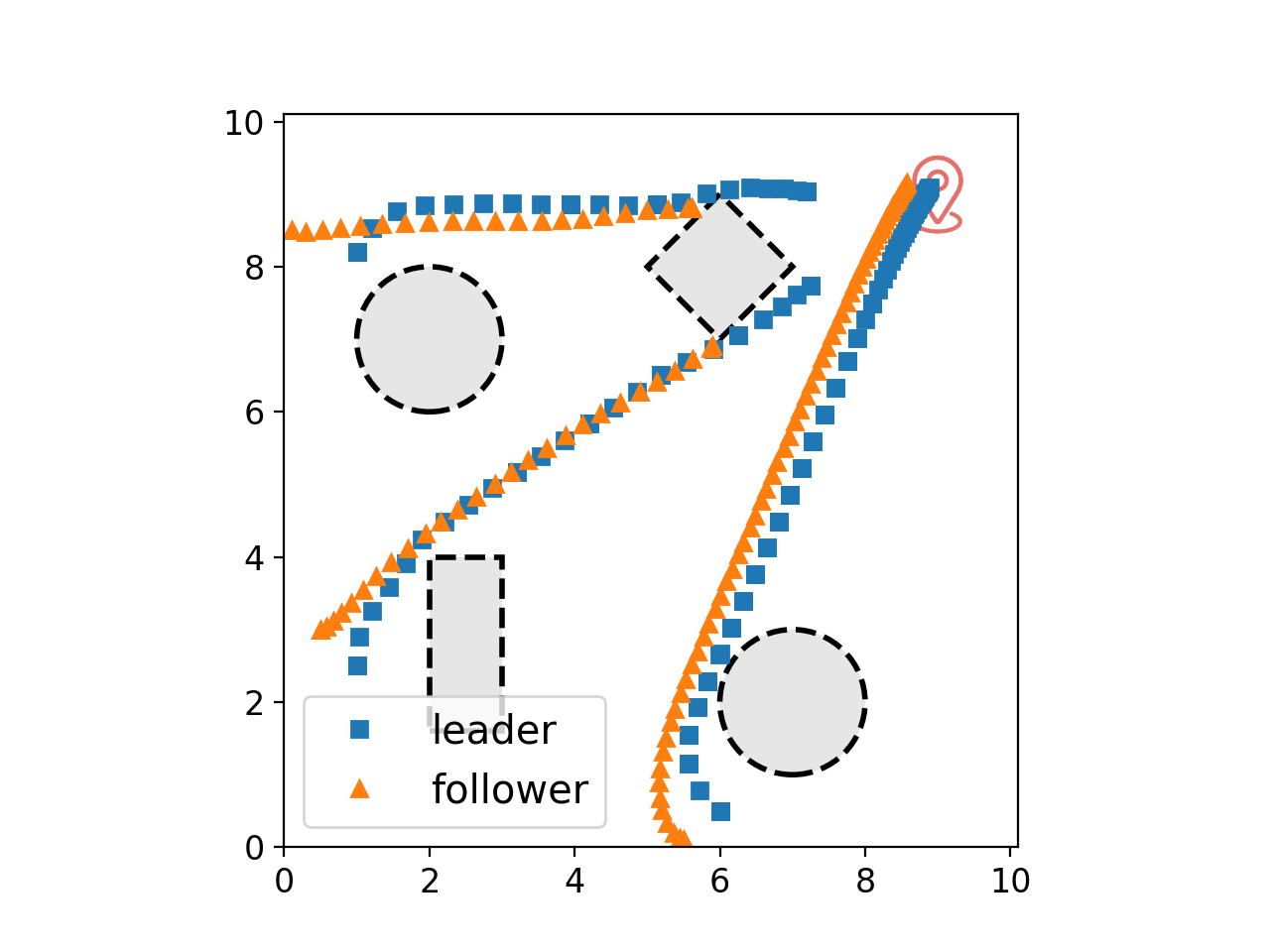}
        \caption{\texttt{dmd}-approach} 
        \label{fig:rc.dmd}
    \end{subfigure}
    \captionsetup{belowskip=-10pt}
    \caption{Interactive guidance trajectories using receding horizon planning and the models learned by different approaches. The blue and the orange represent the leader and follower trajectories, respectively. The follower start from $[0,8.5]$, $[0.5,3]$, and $[5.5,0]$ to reach the destination $[9,9]$. The leader successfully guides the follower using our approach. \texttt{nn}-approach (and \texttt{dmd}-approach) fail to guide the follower in one (and two) case. }
    \label{fig:rc}
\end{figure*}

We choose a $[0,10] \times [0,10]$ working space $\mW$ with 4 obstacles shown in Fig.~\ref{fig:rc}. A discrete unicycle model is used as the leader and follower's dynamical systems:
\begin{equation*}
    x^i_{t+1} = x^i_t + \begin{bmatrix} v^i_t \cos(\theta^i_t) \\ v^i_t \sin(\theta^i_t) \\ \omega^i_t \end{bmatrix} \dd t, \quad i \in \{L, F\}.
\end{equation*}
The state $x^i = [p^i_x \ p^i_y \ \theta^i] \in \R^3$ denotes the ($x,y$)-positions and the heading angle of the robot $i$; the control $u^i = [v^i \ \omega^i] \in [0,2]\times [-2,2]$ represents the linear and angular velocities. The leader's planning horizon is set as $T=5$ and $\dd t = 0.2$. The target destination is $[9,9]$. 

The leader's stage cost is given by
\begin{equation*}
    g^L_t = \| x^L_t - x^F_t \|^2_{Q^L_1} + \| x^L_t - x^\dd \|^2_{Q^L_2} + \| u^L_t \|^2_{R^L},
\end{equation*}
where the first term measures the guidance effectiveness and the second term drives the leader to the destination. The terminal cost of $g^L_T$ is the same as the first two terms of $g^L_t$. We set $Q^L_1 = \diag(2,2,0)$, $R^L = \diag(2,1)$. To implement scenario-based guidance, we set the threshold $\lambda = 1$. When $\lambda < 1$, $Q^L_2 = \diag(1,1,0)$. Otherwise, $Q^L_2 = \diag(0.1, 0.1, 0)$. The purpose is to make the leader approach the follower because greater distance does not facilitate effective guidance.
To simulate interactive data, we use the following \emph{ground truth} cost function as the follower's objective:
\begin{equation*}
\begin{split}
    J^F_t&(u^F_t; x^L_t, u^L_t) = \| x^F_{t+1} - x^L_{t+1} \|^2_{Q^F_1} + \| x^F_{t+1} - x^\dd \|^2_{Q^F_2} \\ 
    &+ Q^F_3 \cdot \tuple{ \Vec{d}^L, \Vec{d}^F } + \| u^F_t \|^2_{R^F}.
\end{split}
\end{equation*}
Here, $\Vec{d}^L = [\cos(\theta^L_{t+1}), \sin(\theta^L_{t+1})] \in \R^2$ and $\Vec{d}^F = [\cos(\theta^F_{t+1}), \sin(\theta^F_{t+1})] \in \R^2$ are the leader's and the follower's heading directions, respectively. $x^L_{t+1} = f^L(x^L_t, u^L_t)$. The first two terms are similar to the ones in $g^L_t$. The third term aligns the follower's heading direction with the leader. We use $Q^F_1 = \diag(10,10,0)$, $Q^F_2 = \diag(0.1,0.1,0)$, $Q^F_3 = -1$, $R^F = \diag(2,0.05)$.

The follower's best response (the inner-level problem) is solved by grid search to avoid local optimal solutions. We collect interactive trajectories with length $S=30$ to generate the data set $\mD$.

\subsection{Koopman Learning Results}
We use a neural network (NN) with three layers of 90 ReLU nonlinearities to parameterize the embedding function $g_x$ in Sec.~\ref{sec:kp.learn} and matrices $\tuple{A,B}$ for the Koopman operator. The training objective is set by a $30$-step prediction loss. 
We perform the training on different sizes of source data. For each size, we conduct the training for $10$ times to evaluate the average learning performance. The model is trained over $10^3$ epochs with SGD each time. All learning is performed on an RTX 6000 GPU, and Tab.~\ref{tab:1} summarizes the training results using different numbers of data. We can observe that the training is stable as the number of samples increases, meaning that the data set size has less impact on learning the Koopman operator.

\begin{figure}
    \centering
    \begin{subfigure}[t]{0.23\textwidth}
        \centering
        \includegraphics[height=3cm]{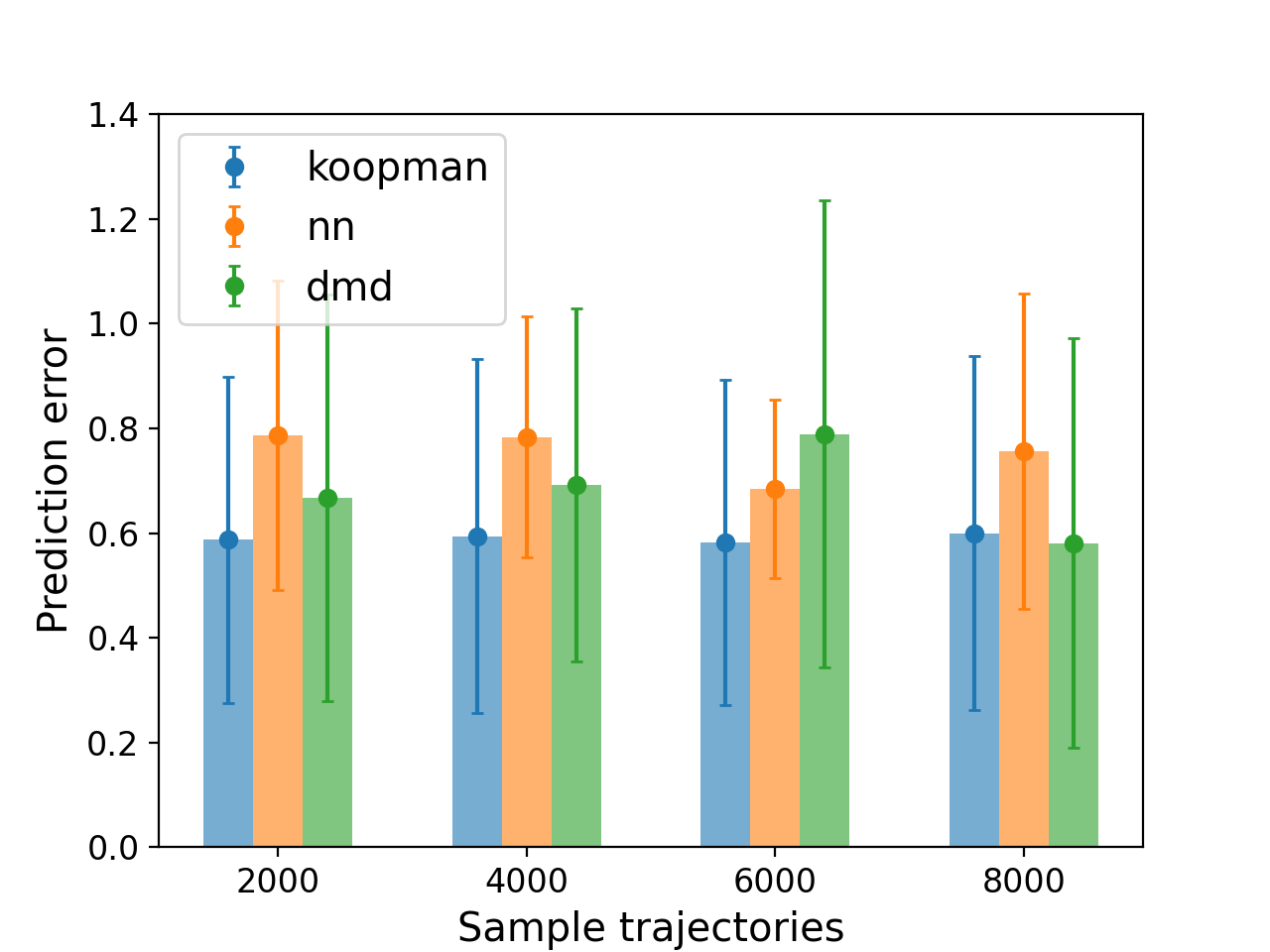}
        \caption{Prediction error at the 10th step over 20 trajectories.}
        \label{fig:err_pred.1}
    \end{subfigure}
    \hspace{2mm}
    \begin{subfigure}[t]{0.23\textwidth}
        \centering
        \includegraphics[height=3cm]{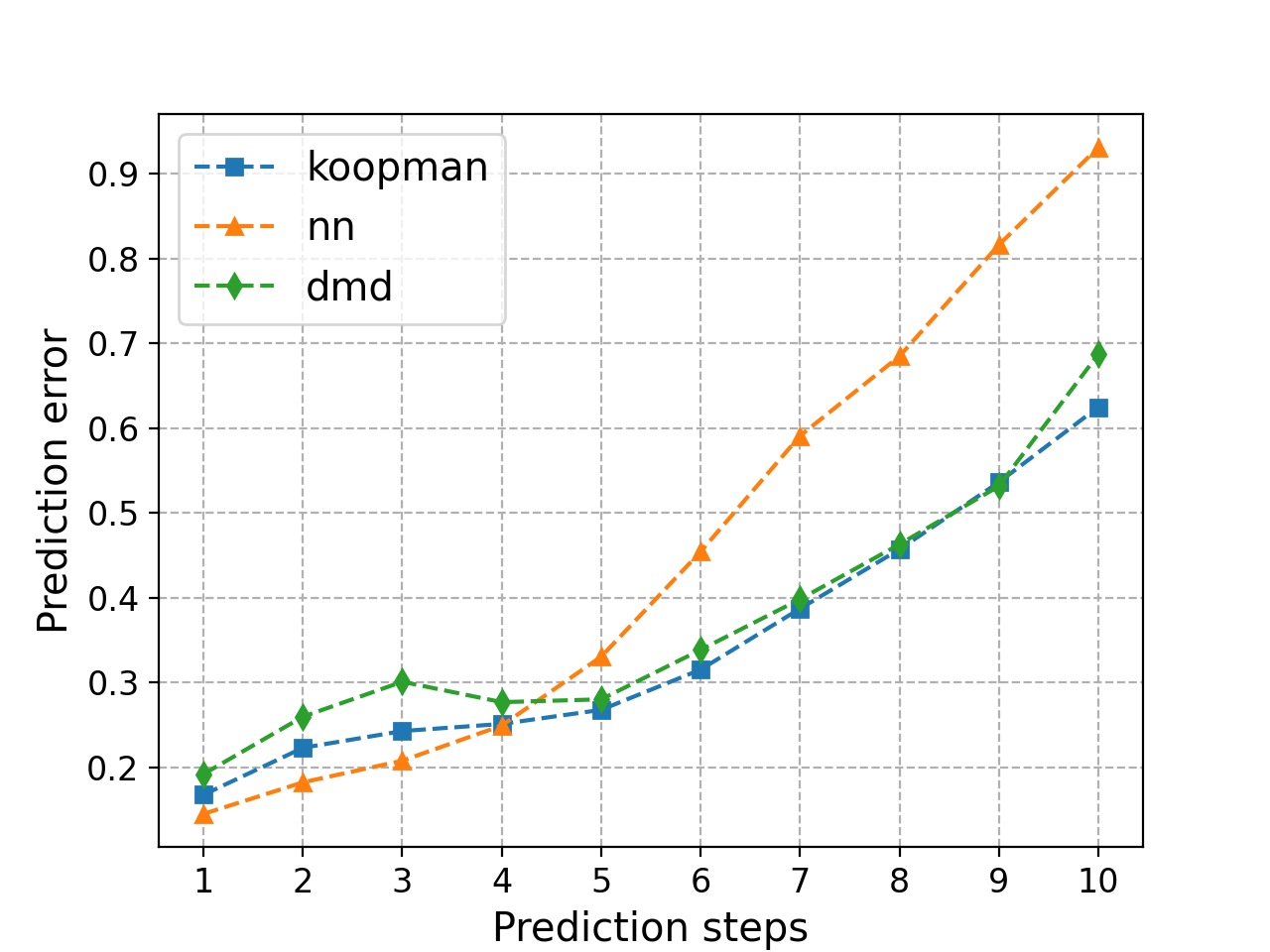}
        \caption{Step-wise prediction error of a single trajectory.}
        \label{fig:err_pred.2}
    \end{subfigure}
    \captionsetup{belowskip=-15pt}
    \caption{Prediction errors with the learned model generated by three approaches. Our Koopman-based approach provides a smaller averaged prediction error and better long-term performance compared with the other two.}
    \label{fig:err_pred}
\end{figure}

\begin{table}[h]
\centering
\begin{tabular}{c|c|c|c|c} \hline
                & $N=2500$  & $N=5000$  & $N=7500$  & $N=10000$ \\ \hline 
    Train loss & 0.178(0.052)  & 0.173(0.045) & 0.172(0.044)  & 0.169(0.042) \\ \hline
    Test loss  & 0.188(0.053)  & 0.228(0.070)  & 0.182(0.045)  & 0.183(0.043) \\ \hline
    Train time  & 18.3 min &  35.4 min &  52.8 min &  70.8 min \\ \hline
\end{tabular}
\caption{Summary of Learning statistics of the Koopman operator. $N$ is the size of the data set, and the training/testing data contains $80/20$. The mean and variance of all losses are averaged over ten training episodes.}
\label{tab:1}
\end{table}

For comparison, we also implement another two methods to learn the follower's feedback dynamics $\tilde{f}^F$. The first one uses an NN of 64 ReLU nonlinearities to fit the one-step dynamics (labeled by  \texttt{nn}-approach). The second one learns a linear model of $\tilde{f}^F$ using Dynamic Mode Decomposition \cite{proctor2016dynamic} (labeled by \texttt{dmd}-approach). They are training with $N=10000$ trajectory samples. 

We use the learned model by three approaches to predict the follower's response trajectory and evaluate the learning performance. The prediction is proceeded by \eqref{eq:kp_linear}. The prediction error is the difference between the predicted and ground truth states. We perform 10-step predictions over 20 ground truth interactive trajectories and plot the last step (the 10th step) prediction error in Fig.~\ref{fig:err_pred.1}. We observe that our approach provides the learned model that generates the smallest prediction error in the first three cases. In the last case with $N=8000$, our approach only generates a slightly larger error compared with \texttt{dmd}-approach. However, the \texttt{dmd}-approach yields a larger variance than ours. 
For better visualization, we plot the step-wise prediction errors of three approaches using one ground truth trajectory in Fig.~\ref{fig:err_pred.2}. It is worth noting that the \texttt{nn}-approach yields a better prediction result in the first three steps. It is because the \texttt{nn}-approach is designed to fit the one-step dynamics directly and hence performs better for short-term prediction. However, it shows poor long-term prediction performance compared with our approach. This is due to the fact that our approach considers the dynamics evolution along trajectories. Also, due to the embedding function, our approach can capture more information in the dynamics compared with \texttt{dmd}-approach and hence has a smaller prediction error. Fig.~\ref{fig:err_pred.2} also suggests that the learned feedback dynamical model by our approach is more suitable for long-term planning.

\subsection{Receding Horizon Planning Results}
After obtaining the trained model, the leader runs Alg.~\ref{alg:1} to achieve guidance. We simulate interactive guidance trajectories from different initial positions using three approaches. Besides, we also simulate baseline trajectories using the model-based OCP \eqref{eq:ocp.foc} to compare our approach, as shown in Fig.~\ref{fig:rc}.

Our approach successfully guides the follower to the destination from three starting points. Compared with the baseline, the follower's terminal position slightly deviates from the exact target $[9,9]$ but is still within an acceptable range, which shows the effectiveness of our approach. This is mainly because of approximation accuracy in the learned model.
However, the other two methods fail the guidance task in some cases. The \texttt{nn}-approach fails to overcome the diamond obstacle starting from the position $[0.5, 3]$ because the leader's prediction on the follower has already diverged near the obstacle. The \texttt{dmd}-approach fails to guide the follower to circumvent the diamond obstacle because of the model accuracy. We observe that the leader has already navigated past the obstacle, and her prediction model indicates that the follower will also successfully avoid the obstacle by adhering to her trajectory. The absence of nonlinearity in the follower's model learned through the \texttt{dmd}-approach leads to inadequacy in providing guidance near obstacles.

\begin{figure}
    \centering
    \begin{subfigure}[t]{0.23\textwidth}
        \centering
        \includegraphics[height=3cm]{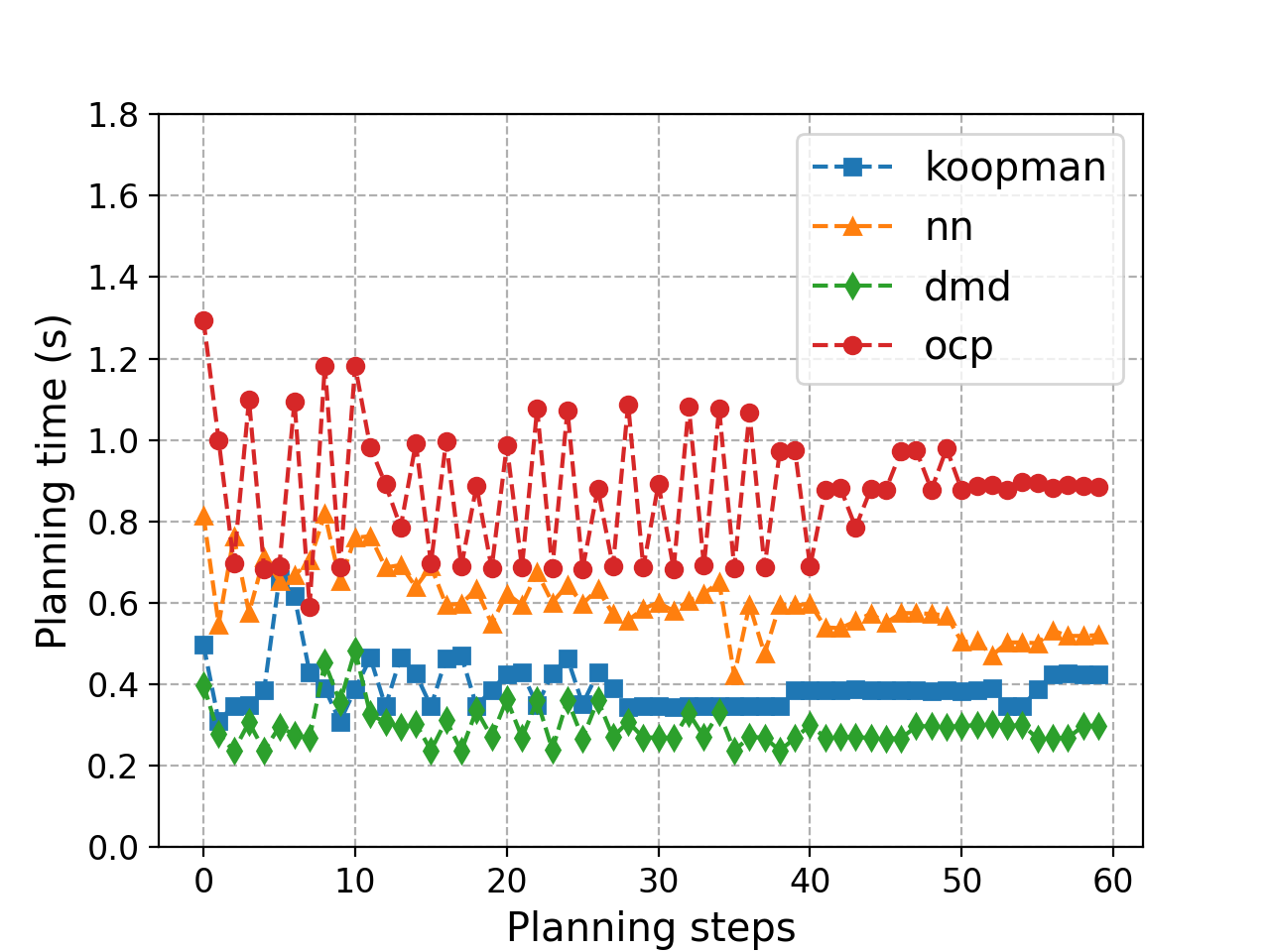}
        \caption{Planning time in each step.}
        \label{fig:rc_plan.time}
    \end{subfigure}
    \hspace{2mm}
    \begin{subfigure}[t]{0.23\textwidth}
        \centering
        \includegraphics[height=3cm]{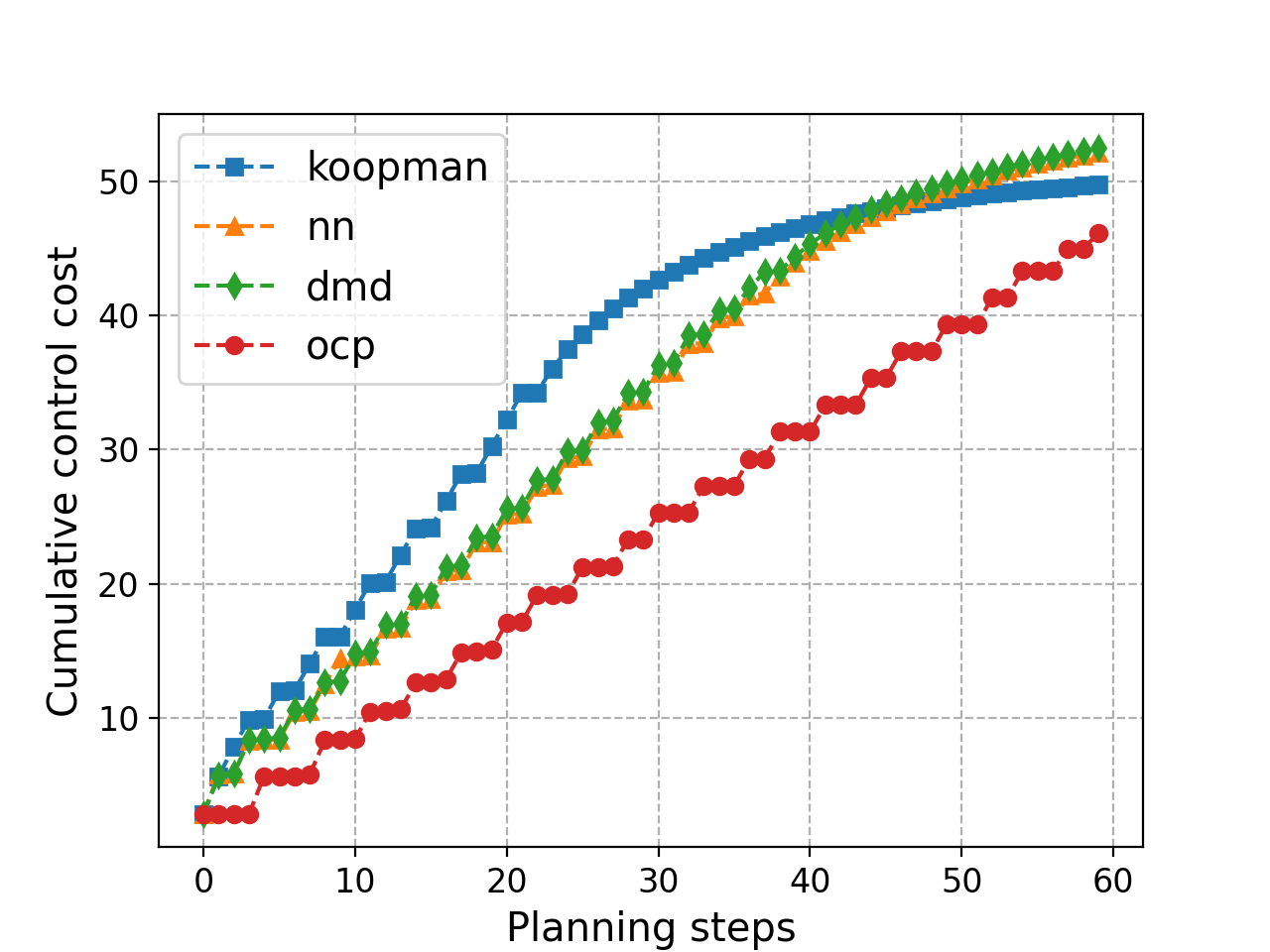}
        \caption{Cumulative control cost.}
        \label{fig:rc_plan.control}
    \end{subfigure}
    \captionsetup{belowskip=-15pt}
    \caption{Planning details for the interactive trajectory starting from $[5.5,0]$.}
    \label{fig:rc_plan}
\end{figure}

We take the guidance trajectory starting from $[5.5,0]$ for detailed discussion since all approaches complete this guidance task. 
We plot the leader's planning time along the trajectory in Fig.~\ref{fig:rc_plan.time}. It is not surprising that the model-based OCP takes the longest time for planning. We note that our approach reduces planning time by about half compared with the modeled-based OCP baseline and still completes the guidance tasks, showing notable time efficiency in guidance planning. The \texttt{dmd}-approach achieves similar planning time as it uses linear models to predict follower dynamics. The \texttt{nn}-based model takes more time to plan as the learned model is nonlinear (an NN). 

We also plot the leader's cumulative control cost in Fig.~\ref{fig:rc_plan.control} to measure the guidance efficiency. The model-based OCP baseline yields the least control effort due to the model knowledge. Our approach provides a higher control cost at the beginning and soon becomes stable, meaning that the guidance is completed in a shorter time. In comparison, \texttt{dmd}-approach and \texttt{nn}-approach shows a higher guidance cost when approaching the destination. It indicates that our approach provides comparably efficient guidance regarding the total control cost.

\section{Conclusions}
In this work, we have proposed a Stackelberg game-theoretic approach to address the challenge of guided trajectory planning when dealing with an unknown follower robot. Our approach not only formulates a game-theoretic framework for multi-robot cooperation in trajectory guidance but also develops an algorithm based on the Koopman operator. This algorithm effectively learns the follower's decision-making model and achieves both rapid and control-efficient guidance. Simulations have corroborated our approach, demonstrating its superior performance in generating a more accurate linear model for predicting the follower's long-term behavior compared to alternative learning methods. Benchmarked with the model-based baseline method, our approach greatly reduces the planning time while successfully accomplishing the guidance task.

There are remaining challenges to ensure safety when employing the Koopman operator for mission-critical applications.  A potential avenue for future research involves an in-depth exploration of safety assurances, encompassing critical aspects such as the establishment of error bounds and the identification of valid operational regions.








\bibliographystyle{IEEEtran}
\bibliography{IEEEabrv,mybib}

\end{document}